\pdfoutput=1

\documentclass[11pt, dvipsnames]{article}

\usepackage[table]{xcolor}
\usepackage[final]{acl}

\usepackage{graphicx}
\usepackage{amsmath}

\usepackage{booktabs}
\usepackage{array,multirow,graphicx}
\usepackage{float}
\usepackage{makecell}
\usepackage{amssymb}
\usepackage{pifont}

\usepackage{algorithm}
\usepackage{algorithmic}
\usepackage{hyperref}
\usepackage{multirow}
\usepackage{multicol}
\usepackage{graphics}
\usepackage{array}
\usepackage{array, diagbox}

\usepackage{booktabs}
\usepackage{arydshln}
\usepackage{multirow}
\usepackage{times}
\usepackage{latexsym}
\usepackage[multiple]{footmisc}

\usepackage{times}
\usepackage{latexsym}

\usepackage[T1]{fontenc}

\usepackage[utf8]{inputenc}

\usepackage{microtype}

\usepackage{inconsolata}

\usepackage{xcolor}

\title{Maverick:\\
Efficient and Accurate Coreference Resolution Defying Recent Trends}

\author{Giuliano Martinelli, Edoardo Barba,
 \normalfont{and} {\bf Roberto Navigli}\\
Sapienza NLP Group, Sapienza University of Rome\\
\texttt{\{martinelli, barba, navigli\}@diag.uniroma1.it}}

\newcommand{\sysns}{Maverick}
\newcommand{\sys}{\sysns~}
\newcommand{\syspipens}{Maverick Pipeline}
\newcommand{\syspipe}{\syspipens~}
\newcommand{\sysstarttoendns}{Maverick$_{\text{s2e}}$}
\newcommand{\sysstarttoend}{Maverick$_{\text{s2e}}$~}
\newcommand{\sysmesns}{Maverick$_{\text{mes}}$}
\newcommand{\sysmes}{Maverick$_{\text{mes}}$~}
\newcommand{\sysincrementalns}{Maverick$_{\text{incr}}$}
\newcommand{\sysincremental}{Maverick$_{\text{incr}}$~}

\newcommand{\abo}{\textcolor{Red}{\textbf{[}}}
\newcommand{\abc}{\textcolor{Red}{\textbf{]$_1$}}}
\newcommand{\bbo}{\textcolor{Green}{\textbf{[}}}
\newcommand{\bbc}{\textcolor{Green}{\textbf{]$_2$}}}
\newcommand{\cbo}{\textcolor{Brown}{\textbf{[}}}
\newcommand{\cbc}{\textcolor{Brown}{\textbf{]$_3$}}}
\newcommand{\dbo}{\textcolor{Blue}{\textbf{[}}}
\newcommand{\dbc}{\textcolor{Blue}{\textbf{]$_4$}}}
\newcommand{\ebo}{\textcolor{Purple}{\textbf{[}}}
\newcommand{\ebc}{\textcolor{Purple}{\textbf{]$_5$}}}
\newcommand{\fbo}{\textcolor{Black}{\textbf{[}}}
\newcommand{\fbc}{\textcolor{Black}{\textbf{]$_6$}}}
\newcommand{\link}{{\small\url{https://github.com/SapienzaNLP/maverick-coref}}}

\begin{document}
\maketitle

\begin{abstract}
Large autoregressive generative models have emerged as the cornerstone for achieving the highest performance across several Natural Language Processing tasks. 
However, the urge to attain superior results has, at times, led to the premature replacement of carefully designed task-specific approaches without exhaustive experimentation. 
The Coreference Resolution task is no exception; all recent state-of-the-art solutions adopt large generative autoregressive models that outperform encoder-based discriminative systems. 
In this work, we challenge this recent trend by introducing \sysns, a carefully designed -- yet simple -- pipeline, which enables running a state-of-the-art Coreference Resolution system within the constraints of an academic budget, outperforming models with up to 13 billion parameters with as few as 500 million parameters.
\sys achieves state-of-the-art performance on the CoNLL-2012 benchmark, training with up to 0.006x the memory resources and obtaining a 170x faster inference compared to previous state-of-the-art systems.
We extensively validate the robustness of the \sys framework with an array of diverse experiments, reporting improvements over prior systems in data-scarce, long-document, and out-of-domain settings.
We release our code and models for research purposes at \link.
\end{abstract}

\section{Introduction}\label{sec:introduction}
As one of the core tasks in Natural Language Processing, Coreference Resolution aims to identify and group expressions (called mentions) that refer to the same entity \cite{karttunen-1969-discourse-referents}. 
Given its crucial role in various downstream tasks, such as Knowledge Graph Construction \cite{li-etal-2020-gaia}, Entity Linking \cite{kundu-etal-2018-neural, agarwal-etal-2022-entity}, Question Answering \cite{dhingra-etal-2018-neural, dasigi-etal-2019-quoref, Bhattacharjee2020InvestigatingQE, chen-durrett-2021-robust}, Machine Translation \cite{stojanovski-fraser-2018-coreference, voita-etal-2018-context, ohtani-etal-2019-context, yehudai-etal-2023-evaluating} and Text Summarization \cite{falke-etal-2017-concept, pasunuru-etal-2021-efficiently, liu-etal-2021-coreference}, \textit{inter alia}, there is a pressing need for both high performance and efficiency.
However, recent works in Coreference Resolution either explore methods to obtain reasonable performance optimizing time and memory efficiency \cite{kirstain-etal-2021-coreference, dobrovolskii-2021-word, otmazgin-etal-2022-f}, or strive to improve benchmark scores regardless of the increased computational demand \cite{bohnet-etal-2023-coreference, zhang2023seq2seq}.

Efficient solutions usually rely on discriminative formulations, frequently employing the mention-antecedent classification method proposed by \citet{lee-etal-2017-end}.
These approaches leverage relatively small encoder-only transformer architectures \cite{joshi-etal-2020-spanbert, beltagy2020longformer} to encode documents and build on top of them task-specific networks that ensure high speed and efficiency.
On the other hand, performance-centered solutions are nowadays dominated by general-purpose large Sequence-to-Sequence models \cite{liu-etal-2022-autoregressive, zhang2023seq2seq}.
A notable example of this formulation, and currently the state of the art in Coreference Resolution, is \citet{bohnet-etal-2023-coreference}, which proposes a transition-based system that incrementally builds clusters of mentions by generating coreference links sentence by sentence in an autoregressive fashion. 
Although Sequence-to-Sequence solutions achieve remarkable performance, their autoregressive nature and the size of the underlying language models (up to 13B parameters) make them dramatically slower and memory-demanding compared to traditional encoder-only approaches. 
This not only makes their usage for downstream applications impractical, but also poses a significant barrier to their accessibility for a large number of users operating within an academic budget.

In this work we argue that discriminative encoder-only approaches for Coreference Resolution have not yet expressed their full potential and have been discarded too early in the urge to achieve state-of-the-art performance.
In proposing  \sysns, we strike an optimal balance between high performance and efficiency, a combination that was missing in previous systems.
Our framework enables an encoder-only model to achieve top-tier performance while keeping the overall model size less than one-twentieth of the current state-of-the-art system, and training it with academic resources.
Moreover, when further reducing the size of the underlying transformer encoder, \sys performs in the same ballpark as encoder-only efficiency-driven solutions while improving speed and memory consumption.
Finally, we propose a novel incremental Coreference Resolution method that, integrated into the \sys framework, results in a robust architecture for out-of-domain, data-scarce, and long-document settings.

\section{Related Work}\label{sec:related-works}
We now introduce well-established approaches to neural Coreference Resolution.
Specifically, we first delve into the details of traditional discriminative solutions, including their incremental variations, and then present the recent paradigm shift for approaches based on large generative architectures.

\subsection{Discriminative models}\label{sec:related-works-discriminative}
Discriminative approaches tackle the Coreference Resolution task as a classification problem, usually employing encoder-only architectures.
The pioneering works of \citet{lee-etal-2017-end, lee-etal-2018-higher} introduced the first end-to-end discriminative system for Coreference Resolution, the Coarse-to-Fine model.
First, it involves a mention extraction step, in which the spans most likely to be coreference mentions are identified.
This is followed by a mention-antecedent classification step where, for each extracted mention, the model searches for its most probable antecedent (i.e., the extracted span that appears before in the text).
This pipeline, composed of mention extraction and mention-antecedent classification steps, has been adopted with minor modifications in many subsequent works, that we refer to as \textit{Coarse-to-Fine} models.

\paragraph{Coarse-to-Fine Models}
Among the works that build upon the Coarse-to-Fine formulation, \citet{lee-etal-2018-higher}, \citet{joshi-etal-2019-bert} and \citet{joshi-etal-2020-spanbert} experimented with changing the underlying document encoder,  utilizing ELMo \cite{peters-etal-2018-deep}, BERT \cite{devlin-etal-2019-bert} and SpanBERT \cite{joshi-etal-2020-spanbert}, respectively, achieving remarkable score improvements on the English OntoNotes \cite{pradhan-etal-2012-conll}. 
Similarly, \citet{kirstain-etal-2021-coreference} introduced s2e-coref that reduces the high memory footprint of SpanBERT by leveraging the LongFormer \cite{beltagy2020longformer} sparse-attention mechanism. 
Based on the same architecture, \citet{otmazgin-etal-2023-lingmess} analyzed the impact of having multiple experts score different linguistically motivated categories (e.g., pronouns-nouns, nouns-nouns, etc.).
While the foregoing works have been able to modernize the original Coarse-to-Fine formulation, training their architectures on the OntoNotes dataset still requires a considerable amount of memory.\footnote{Training those models requires at least 32G of VRAM.}
This occurs because they rely on the traditional Coarse-to-Fine pipeline that, as we cover in Section \ref{sec:methodology-sys-pipeline}, has a large memory overhead and is based on manually-set thresholds to regulate memory usage. 

\paragraph{Incremental Models}
Discriminative systems also include incremental techniques. 
Incremental Coreference Resolution has a strong cognitive grounding: research on the “garden-path” effect shows that humans resolve referring expressions incrementally \cite{ALTMANN1988191}.

A seminal work that proposed an automatic incremental system was that of \citet{webster-curran-2014-limited}, which introduced a clustering approach based on the shift-reduce paradigm. In this formulation, for each mention, a classifier decides whether to SHIFT it into a singleton (i.e., single mention cluster) or to REDUCE it within an existing cluster.
The same approach has recently been reintroduced in ICoref \cite{xia-etal-2020-incremental} and longdoc \cite{toshniwal-etal-2021-generalization}, which adopted SpanBERT and LongFormer, respectively. 
In these works the mention extraction step is identical to that of Coarse-to-Fine models. 
On the other hand, the mention clustering step is performed by using a linear classifier that scores each mention against a vector representation of previously built clusters, in an incremental fashion. 
This method ensures constant memory usage since cluster representations are updated with a learnable function.
In Section \ref{sec:methodology-sys-clustering} we present a novel performance-driven incremental method that obtains superior performance and generalization capabilities, in which we adopt a lightweight transformer architecture that retains the mention representations.

\subsection{Sequence-to-Sequence models}\label{sec:related-works-sequence}
Recent state-of-the-art Coreference Resolution systems all employ autoregressive generative approaches. 
However, an early example of Sequence-to-Sequence model, TANL \cite{paolini2021structured}, failed to achieve competitive performance on OntoNotes.
The first system to show that the autoregressive formulation was competitive was ASP \cite{liu-etal-2022-autoregressive}, which outperformed encoder-only discriminative approaches.
ASP is an autoregressive pointer-based model that generates actions for mention extraction (bracket pairing) and then conditions the next step to generate coreference links.
Notably, the breakthrough achieved by ASP is not only due to its formulation but also to its usage of large generative models. 
Indeed, the success of their approach is strictly correlated with the underlying model size, since, when using models with a comparable number of parameters, the performance is significantly lower than encoder-only approaches.
The same occurs in \citet{zhang2023seq2seq}, a fully-seq2seq approach where a model learns to generate a formatted sequence encoding coreference notation, in which they report a strong positive correlation between performance and model sizes.

Finally, the current state-of-the-art system on the OntoNotes benchmark is held by Link-Append \cite{bohnet-etal-2023-coreference}, a transition-based system that incrementally builds clusters exploiting a multi-pass Sequence-to-Sequence architecture. 
This approach incrementally maps the mentions in previously coreference-annotated sentences to system actions for the current sentence, using the same shift-reduce incremental paradigm presented in Section \ref{sec:related-works-discriminative}. 
This method obtains state-of-the-art performance at the cost of using a 13B-parameter model and processing one sentence at a time, drastically increasing the need for computational power.
While the foregoing models ensure superior performance compared to previous discriminative approaches, using them for inference is out of reach for many users, not to mention the exorbitant cost of training them from scratch.

\section{Methodology}\label{sec:methodology}
In this section, we present the \sys framework: we propose replacing the preprocessing and training strategy of Coarse-to-Fine models with a novel pipeline that improves the training and inference efficiency of Coreference Resolution systems.
Furthermore, with the \syspipens, we eliminate the dependency on long-standing manually-set hyperparameters that regulate memory usage.
Finally, building on top of our pipeline, we propose three models that adopt a mention-antecedent classification technique, namely \sysstarttoend and \sysmesns, and a system that is based upon a novel incremental formulation, \sysincrementalns.

\subsection{\sys Pipeline}\label{sec:methodology-sys-pipeline}
The \syspipe combines i) a novel mention extraction method, ii) an efficient mention regularization technique, and iii) a new mention pruning strategy.

\paragraph{Mention Extraction}
When it comes to extracting mentions from a document $D$, there are different strategies to model the probability that a span contains a mention.
Several previous works follow the Coarse-to-Fine formulation presented in Section \ref{sec:related-works-discriminative}, which consists of scoring all the possible spans in $D$. 
This entails a quadratic computational cost in relation to the input length, which they mitigate by introducing several pruning techniques.

In this work, we employ a different strategy.
We extract coreference mentions by first identifying all the possible starts of a mention, and then, for each start, extracting its possible end.
To extract start indices, we first compute the hidden representation $(x_1, \dots, x_n)$ of  the tokens $(t_1, \dots, t_n) \in D$ using a transformer encoder, and then use a fully-connected layer $F$ to compute the probability for each $t_i$ being the start of a mention as:
\begin{equation*}
    F_{start}(x) = W_{start}'(GeLU(W_{start}x)) 
\end{equation*}
\begin{equation*}
    p_{start}(t_i) = \sigma(F_{start}(x_i))
\end{equation*}

\noindent with $W'_{start}, W_{start}$ being the learnable parameters, and $\sigma$ the sigmoid function.
For each start of a mention $t_s$, i.e., those tokens having $p_{start}(t_s) > 0.5$, we then compute the probability of its subsequent tokens $t_j$, with $ s \leq j$, to be the end of a mention that starts with $t_s$.
We follow the same process as that of the mention start classification, but we condition the prediction on the starting token by concatenating the start, $x_s$, and end, $x_j$, hidden representations before the linear classifier:
\begin{equation*}
F_{end}(x, x') = W'_{end}(GeLU(W_{end}[x, x']))
\end{equation*}
\begin{equation*}
p_{end}(t_j | t_s) = \sigma(F_{end}(x_s, x_j))
\end{equation*}
\noindent with $W'_{end}$, $W_{end}$ being learnable parameters.
This formulation handles overlapping mentions since, for each start $t_s$, we can find multiple ends $t_e$ (i.e., those that have $p_{end}(t_j|t_s) > 0.5$).

Previous works already adopted a linear layer to compute start and end mention scores for each possible mention, i.e., s2e-coref \cite{kirstain-etal-2021-coreference}, and LingMess \cite{otmazgin-etal-2023-lingmess}.
However, our mention extraction technique differs from previous approaches since i) we produce two probabilities $(0 < p < 1)$ instead of two unbounded scores and ii) we use the computed start probability to filter out possible mentions, which reduces by a factor of 9 the number of mentions considered compared to existing Coarse-to-Fine systems (Table \ref{table:syspipe}, first row).

\paragraph{Mention Regularization}
To further reduce the computation demand of this process, in the \syspipe we use the end-of-sentence (EOS) mention regularization strategy: 
after extracting the span start, we consider only the tokens up to the nearest EOS as possible mention end candidates.\footnote{We note that all the well-established Coreference Resolution datasets are sentence-split.}
Since annotated mentions never span across sentences, EOS mention regularization prunes the number of mentions considered without any loss of information.
While this heuristic was initially introduced in the implementation of \citet{lee-etal-2018-higher}, all the recent Coarse-to-Fine have abandoned it in favor of the maximum span-length regularization, which is a manually-set hyperparameter that regulates a threshold to filter out spans that exceed a certain length.
This implies a large overhead of unnecessary computations and introduces a structural bias that does not consider long mentions that exceed a fixed length.\footnote{The max-length regularization filters out 196 correctly annotated spans when training on OntoNotes.}
In our work, we not only reintroduce the EOS mention regularization, but we also study its contribution in terms of efficiency, as reported in Table \ref{table:syspipe}, second row.

\paragraph{Mention Pruning} After the mention extraction step, as a result of the \syspipens, we consider an 18x lower number of candidate mentions for the successive mention clustering phase (Table \ref{table:syspipe}). 
This step consists of computing, for each mention, the probability of all its antecedents being in the same cluster, incurring a quadratic computational cost.
Within the Coarse-to-Fine formulation, this high computational cost is mitigated by considering only the top $k$ mentions according to their probability score, where $k$ is a manually set hyperparameter.
Since after our mention extraction step we obtain probabilities for a very concise number of mentions, we consider only mentions classified as probable candidates (i.e., those with $p_{end} > 0.5$ and $p_{start} > 0.5$),  reducing the number of mention pairs considered by a factor of 10.
In Table \ref{table:syspipe}, we compare the previous Coarse-to-Fine formulation with the new \syspipens.

\begin{table}[t]
\centering
\resizebox{\columnwidth}{!}{
\begin{tabular}{l|c|c|c}
\hline
& Coarse-to-Fine & \sys & $\Delta$ \\
 \hline
 \hline
 Ment. Extraction & Enumeration & (i) Start-End & \\
& 183,577 & 20,565 & -8,92x\\
(+) Regularization & (+) Span-length  & (ii) (+) EOS & \\ 
& 14,265 &  777 & -18,3x\\ 
 \hline
Ment. Clustering & Top-k & (iii) Pred-only &  \\  
 & 29,334 &  2,713 & -10,81x\\
\hline
\end{tabular}
}
\caption{
Comparison between the Coarse-to-Fine pipeline and the \syspipe in terms of the average number of mentions considered in the mention extraction step (top) and the average number of mention pairs considered in the mention clustering step (bottom). The statistics are computed on the OntoNotes devset, and refer to the hyperparameters proposed in \citet{lee-etal-2018-higher}, which were unchanged by subsequent Coarse-to-Fine works, i.e., span-len = 30, top-k = 0.4.}
\label{table:syspipe}
\end{table}

\subsection{Mention Clustering}\label{sec:methodology-sys-clustering}
As a result of the \syspipens, we obtain a set of candidate mentions $M = (m_1, m_2, \dots, m_l)$, for which we propose three different clustering techniques:
\sysstarttoend and \sysmesns, which use two well-established Coarse-to-Fine mention-antecedent techniques, and \sysincrementalns, which adopts a novel incremental technique that leverages a light transformer architecture.

\paragraph{Mention-Antecedent models} 
The first proposed model, \sysstarttoendns, adopts an equivalent mention clustering strategy to \citet{kirstain-etal-2021-coreference}:
given a mention $m_i = (x_s, x_e)$ and its antecedent $m_j=(x_{s'}, x_{e'})$, with their start and end token hidden states, we use two fully-connected layers to model their corresponding representations: 
 \begin{equation*}F_s(x) = W_{s}'(GeLU(W_{s}x))
\end{equation*}
 \begin{equation*}F_e(x) = W_{e}'(GeLU(W_{e}x))
\end{equation*}
\noindent we then calculate their probability to be in the same cluster as:
\begin{align*}
p_c(m_i, m_j) = \sigma (F_s(x_s) \cdot W_{ss} \cdot  F_s(x_{s'}) + \hspace{10mm}\\
F_e(x_e) \cdot W_{ee} \cdot F_e(x_{e'}) +  \hspace{10mm}\\
F_s(x_s) \cdot W_{se} \cdot F_e(x_{e'}) + \hspace{10mm} \\ 
F_e(x_e) \cdot W_{es} \cdot F_s(x_{s'}))\hspace{12mm}
\end{align*}
\noindent with $W_{ss}, W_{ee}, W_{se}, W_{es}$ being four learnable matrices and $W_{s}, W_{s}', W_{e}, W_{e}'$ the learnable parameters of the two fully-connected layers.

A similar formulation is adopted in \sysmesns, where, instead of using only one generic mention-pair scorer, we use 6 different scorers that handle linguistically motivated categories, as introduced by \citet{otmazgin-etal-2023-lingmess}.
We detect which category $k$ a pair of mentions $m_i$ and $m_j$ belongs to (e.g., if $m_i$ is a pronoun and $m_j$ is a proper noun, the category will be \textsc{Pronoun-Entity}) and use a category-specific scorer to compute $p_c$.
A complete description of the process along with the list of categories can be found in Appendix \ref{app:lingmess}.

\paragraph{Incremental model} Finally, we introduce a novel incremental approach to tackle the mention clustering step, namely \sysincrementalns, which follows the standard shift-reduce paradigm introduced in Section \ref{sec:related-works-discriminative}.
Differently from the previous neural incremental techniques (i.e., ICoref \cite{xia-etal-2020-incremental} and longdoc \cite{toshniwal-etal-2021-generalization}) which use a linear classifier to obtain the clustering probability between each mention and a fixed length vector representation of previously built clusters, \sysincremental leverages a lightweight transformer model to attend to previous clusters, for which we retain the mentions' hidden representations. 
Specifically, we compute the hidden representations $(h_1, \dots, h_l)$ for all the candidate mentions in $M$ using a fully-connected layer on top of the concatenation of their start and end token representations.
We first assign the first mention $m_1$ to the first cluster $c_1=(m_1)$.
Then, for each mention $m_i \in M$ at step $i$ we obtain the probability of $m_i$ being in a certain cluster $c_j$ by encoding $h_i$ with all the representations of the mentions contained in the cluster $c_j$ using a transformer architecture.
We use the first special token ([CLS]) of a single-layer transformer architecture $T$ to obtain the score $S(m_i, c_j)$ of $m_i$ being in the cluster $c_j=(m_f, \dots, m_g)$ with $f \leq g < i$ as:
 \begin{equation*}S(m_i, c_j) = W_c \cdot (ReLU(T_{CLS}(h_i, h_f, \dots, h_g)))
\end{equation*}
Finally, we compute the probability of $m_i$ belonging to $c_j$ as:
 \begin{equation*}p_c(m_i \in c_j| c_j = (m_f, \dots, m_g)) = \sigma(S(m_i, c_j)) 
\end{equation*}
 We calculate this probability for each cluster $c_j$ up to step $i$.
 We assign the mention $m_i$ to the most probable cluster $c_j$ having $p_c(m_i \in c_j) > 0.5$ if one exists, or we create a new singleton cluster containing $m_i$.


As we show in Sections \ref{sec:results-ood-evaluation} and \ref{sec:results-comparison-discriminative-models}, this formulation obtains better results than previous incremental methods, and is beneficial when dealing with long-document and out-of-domain settings.

\subsection{Training}\label{sec:methodology-sys-training}
To train a \sys model, we optimize the sum of three binary cross-entropy losses: 
 \begin{equation*}L_{coref} = L_{start} + L_{end} + L_{clust} 
\end{equation*}
Our loss formulation differs from previous transformer-based Coarse-to-Fine approaches, which adopt the marginal log-likelihood to optimize the mention to antecedent score \cite{lee-etal-2018-higher, kirstain-etal-2021-coreference}.
Since their formulation ``makes learning slow and ineffective, especially for mention detection'' \cite{zhang-etal-2018-neural-coreference}, we directly optimize both mention extraction and mention clustering with a multitask approach. 
$L_{start}$ and $L_{end}$ are the start loss and end loss, respectively, of the mention extraction step, and are defined as:
\begin{align*}
L_{start} = \sum\limits_{i=1}^N -(y_i\log(p_{start}(t_i)) + \hspace{15mm} \\
(1 - y_i)\log(1 - p_{start}(t_i))) \hspace{10mm} \\
L_{end} = \sum\limits_{s=1}^{S}\sum\limits_{j=1}^{E_{s}} -(y_i\log(p_{end}(t_j|t_s)) +  \hspace{8mm}\\
(1 - y_i)\log(1 - p_{end}(t_j|t_s))) \hspace{7mm}
\end{align*}
where $N$ is the sequence length, $S$ is the number of starts, $E_s$ is the number of possible ends for a start $s$ and $p_{start}(t_i)$ and $p_{end}(t_j|t_s)$ are those defined in Section \ref{sec:methodology-sys-pipeline}.

Finally, $L_{clust}$ is the loss for the mention clustering step. Since we experiment with two different mention clustering formulations, we use a different loss for each clustering technique, namely $L_{clust}^{ant}$ for the mention-antecedent models, i.e., \sysstarttoend and \sysmesns,  and $L_{clust}^{incr}$ for the incremental model, i.e.,  \sysincremental: 
\begin{align*}
L_{clust}^{ant} = \sum\limits_{i=1}^{|M|}\sum\limits_{j=1}^{|M|}-(y_i\log(p_{c}(m_i|m_j)) + \hspace{8mm}  \\
(1 - y_i)\log(1 - p_{c}(m_i|m_j))) \hspace{10mm}  \\
L_{clust}^{incr} =\sum\limits_{i=1}^{|M|} (\sum\limits_{j=1}^{C_{i}}-(y_i\log(p_{c}(m_i \in c_j)) + \hspace{5mm} \\ 
(1 - y_i)\log(1 - p_{c}(m_i \in c_j)))) \hspace{10mm} 
\end{align*}
where $|M|$ is the number of extracted mentions, $C_i$ is the set of clusters created up to step $i$, and  $p_{c}(m_i|m_j)$ and $p_{c}(m_i \in c_j)$ are  defined in Section \ref{sec:methodology-sys-clustering}.

All the models we introduce are trained using teacher forcing. 
In particular, in the mention token end classification step, we use gold start indices to condition the end tokens prediction, and, for the mention clustering step, we consider only gold mention indices.
For \sysincrementalns, at each iteration, we compare each mention only to previous gold clusters.

\begin{table}[t]
\centering
\resizebox{\columnwidth}{!}{
\begin{tabular}{l|rrrrrr}
\hline
\multicolumn{1}{l}{Dataset} & \multicolumn{1}{r}{\# Train}  & \multicolumn{1}{r}{\# Dev} & \multicolumn{1}{r}{\# Test} & \multicolumn{1}{r}{Tokens} & \multicolumn{1}{r}{Mentions} & \multicolumn{1}{r}{ \% Sing} \\ 
 \hline
OntoNotes  & 2802 & 343 & 348 & 467 & 56 & 0 \\
LitBank & 80 & 10 & 10 & 2105 & 291 & 19.8\\ 
PreCo & 36120 &  500 & 500 & 337 & 105 & 52.0 \\ 
GAP & - & - & 2000 & 95 & 3 & - \\ 
WikiCoref & - & - & 30 & 1996 & 230 & 0 \\  
\hline
\end{tabular}
}
\caption{Dataset statistics: number of documents in each dataset split, average number of words and mentions per document, and singletons percentage.}
\label{table:dataset}
\end{table}

\section{Experiments Setup}\label{sec:setup}
\subsection{Datasets}\label{sec:setup-datasets}
We train and evaluate all the comparison systems on three Coreference Resolution datasets: 

\paragraph{OntoNotes} \cite{pradhan-etal-2012-conll}, proposed in the CoNLL-2012 shared task, is the de facto standard dataset used to benchmark Coreference Resolution systems. 
It consists of documents that span seven distinct genres, including full-length documents (broadcast news, newswire, magazines, weblogs, and Testaments) and multiple speaker transcripts (broadcast and telephone conversations).

\paragraph{LitBank} \cite{bamman-etal-2020-annotated} contains $100$ literary documents typically used to evaluate long-document Coreference Resolution. 

\paragraph{PreCo} \cite{chen-etal-2018-PreCo} is a large-scale dataset that includes reading comprehension tests for middle school and high school students.

\medbreak
Notably, both LitBank and PreCo have different annotation guidelines compared to OntoNotes, and provide annotation for singletons (i.e., single-mention clusters).
Furthermore, we evaluate models trained on OntoNotes on three out-of-domain datasets: 
\vspace{-1mm}
\begin{itemize}
    \item \textbf{GAP} \cite{webster-etal-2018-mind} contains sentences in which, given a pronoun, the model has to choose between two candidate mentions.
    \item \textbf{LitBank$_{\text{ns}}$} and \textbf{PreCo$_{\text{ns}}$}, the datasets' test-set where we filter out singleton annotations.
    \item \textbf{WikiCoref} \cite{ghaddar-langlais-2016-wikicoref}, which contains Wikipedia texts, including documents with up to 9,869 tokens.  
\vspace{-1mm}
\end{itemize}
The statistics of the datasets used are shown in Table \ref{table:dataset}.

\subsection{Comparison Systems}\label{sec:setup-comparison-systems}
\paragraph{Discriminative}
Among the discriminative systems, we consider c2f-coref \cite{joshi-etal-2020-spanbert} and s2e-coref \cite{ kirstain-etal-2021-coreference}, which build upon the Coarse-to-Fine formulation and adopt different document encoders.
We also report the results of LingMess \cite{otmazgin-etal-2023-lingmess}, which is the previous best encoder-only solution, and f-coref \cite{otmazgin-etal-2022-f}, which is a distilled version of LingMess.
Furthermore, we include CorefQA \cite{wu-etal-2020-corefqa}, which casts Coreference as extractive Question Answering, and wl-coref \cite{dobrovolskii-2021-word}, which first predicts coreference links between words, then extracts mentions spans. 
Finally, we report the results of incremental systems, such as ICoref \cite{xia-etal-2020-incremental} and longdoc \cite{toshniwal-etal-2021-generalization}.

\paragraph{Sequence-to-Sequence}
We compare our models with TANL \cite{paolini2021structured} and ASP \cite{liu-etal-2022-autoregressive}, which frame Coreference Resolution as an autoregressive structured prediction.
We also include Link-Append \cite{bohnet-etal-2023-coreference}, a transition-based system that builds clusters with a multi-pass Sequence-to-Sequence architecture. 
Finally, we report the results of seq2seq \cite{zhang2023seq2seq}, a model that learns to generate a sequence with Coreference Resolution labels.

\subsection{\sys Setup}\label{sec:setup-sys}
All \sys models use DeBERTa-v3 \cite{he2023debertav3} as the document encoder.
We use DeBERTa because it can model very long input texts effectively \cite{he2021deberta}.\footnote{This is because its attention mechanism enables its input length to grow linearly with the number of its layers.}
Moreover, compared to the LongFormer, which was previously adopted by several token-level systems, DeBERTa ensures a larger input max sequence length (e.g., DeBERTa$_{\text{large}}$  can handle sequences up to 24,528 tokens while LongFormer only 4096) and has shown better performances empirically in our experiments on the OntoNotes dataset.
On the other hand, using DeBERTa to encode long documents is computationally expensive because its attention mechanism incurs a quadratic computational complexity.
Whereas this further increases the computational cost of traditional Coarse-to-Fine systems, the \syspipe enables us to train models that leverage DeBERTa$_{\text{large}}$ on the OntoNotes dataset, without any performance-lowering pruning heuristic.
To train our models we use Adafactor \cite{pmlr-v80-shazeer18a} as our optimizer, with a learning rate of 3e-4 for the linear layers, and 2e-5 for the pre-trained encoder. 
We perform all our experiments within an academic budget, i.e., a single RTX 4090, which has 24GB of VRAM.
We report more training details in Appendix \ref{app:training}.

\section{Results}\label{sec:results}

\begin{table*}[t!]
\centering
\small
\resizebox{\textwidth}{!}{
\begin{tabular}{l|l|c|r|cr|rc}
\hline
\multicolumn{1}{c}{Model}  & \multicolumn{1}{c}{LM} & \multicolumn{1}{l}{Avg. F1} & \multicolumn{1}{c}{Params} & \multicolumn{2}{c}{Training} & \multicolumn{2}{c}{Inference}\\
\hline
 & & & & Time & Hardware & Time  & Mem.\\
\rowcolor{gray!20}
 \hline
\multicolumn{8}{c}{\textbf{Discriminative}} \\
c2f-coref \cite{joshi-etal-2020-spanbert} & SpanBERT$_{\text{large}}$ & 79.6 & 370M & - & 1x32G & 50s & 11.9~~ \\
ICoref \cite{xia-etal-2020-incremental} & SpanBERT$_{\text{large}}$ & 79.4 & 377M & 40h& 1x1080TI-12G  & 38s & 2.9  \\
CorefQA \cite{wu-etal-2020-corefqa} & SpanBERT$_{\text{large}}$ & ~~83.1* & 740M & -  & 1xTPUv3-128G & - & - \\
s2e-coref 
\cite{kirstain-etal-2021-coreference}
& LongFormer$_{\text{large}}$ & 80.3 & 494M & - & 1x32G & 17s & 3.9\\
longdoc \cite{toshniwal-etal-2021-generalization} &  LongFormer$_{\text{large}}$ & 79.6 & 471M & 16h & 1xA6000-48G & 25s & 2.1 \\
wl-coref \cite{dobrovolskii-2021-word} & RoBERTa$_{\text{large}}$ & 81.0 & 360M & ~~5h & 1xRTX8000-48G & 11s & 2.3\\
f-coref \cite{otmazgin-etal-2022-f} & DistilRoBERTa & ~~78.5* & 91M & - &  1xV100-32G & 3s & 1.0\\
LingMess \cite{otmazgin-etal-2023-lingmess} & LongFormer$_{\text{large}}$ & 81.4 & 590M & 23h & 1xV100-32G & 20s & 4.8\\
\rowcolor{gray!20}
\hline
\multicolumn{8}{c}{\textbf{Sequence-to-Sequence}} \\ 
\multirow{2}{*}{ASP \cite{liu-etal-2022-autoregressive}} 
 & FLAN-T5L & 80.2 & 770M & - & 1xA100-40G & - & - \\
 & FLAN-T5xxl & 82.5 & 11B  & 45h & 6xA100-80G & 20m & -\\ 
\hline
\multirow{2}{*}{Link-Append \cite{bohnet-etal-2023-coreference}}  & mT5xl & ~~78.0$^d$ & 3B  & - & 128xTPUv4-32G & - & - \\  
& mT5xxl & 83.3 & 13B  & 48h & 128xTPUv4-32G & 30m & -\\  
\hline
\multirow{2}{*}{seq2seq \cite{zhang2023seq2seq}} 
& T5-large & ~~77.2$^d$ & 770M & - & 8xA100-40G & - & - \\
 & T0-11B & 83.2 & 11B & - & 8xA100-80G & 40m & - \\ 
 \hline
\rowcolor{gray!20}
\multicolumn{8}{c}{\textbf{Ours (Discriminative)}} \\ 
\multirow{2}{*}{\sysstarttoend }
& DeBERTa$_{\text{base}}$ & 81.1 & 192M & ~~7h & 1xRTX4090-24G & 6s & 1.8\\ 
& DeBERTa$_{\text{large}}$ & 83.4 & 449M & 14h & 1xRTX4090-24G  & 13s & 4.0\\ 
\hline
\multirow{2}{*}{\sysincremental } 
& DeBERTa$_{\text{base}}$  & 81.0 & 197M & 21h & 1xRTX4090-24G & 22s & 1.8 \\
& DeBERTa$_{\text{large}}$  & 83.5 & 452M & 29h & 1xRTX4090-24G & 29s & 3.4\\
\hline
\multirow{2}{*}{\sysmes }  
& DeBERTa$_{\text{base}}$ & 81.4 & 223M & ~~7h & 1xRTX4090-24G & 6s & 1.9 \\ 
& DeBERTa$_{\text{large}}$ & \textbf{83.6} & 504M & 14h & 1xRTX4090-24G & 14s & 4.0\\ 
\hline
\end{tabular}
}
\caption{Results on the OntoNotes benchmark. We report the Avg. CoNLL-F1 score, the number of parameters, the training time, and the hardware used to train each model. Inference time (sec) and memory (GiB) were calculated on an RTX4090. For Sequence-to-Sequence models we include statistics that are reported in the original papers, since we could not run models locally. 
(*) indicates models trained on additional resources. ($^d$) indicates scores obtained on the dev set, however, \sys systems always perform better on the dev than on the test sets. 
Missing values (-) are not reported in the original paper, and it is not feasible to reproduce them using our limited hardware resources.}
\label{table:ontonotes-parames-performance}
\end{table*}

\subsection{English OntoNotes}\label{sec:results-ontonotes}
We report in Table \ref{table:ontonotes-parames-performance} the average CoNLL-F1 score of the comparison systems trained on the English OntoNotes, along with their underlying pre-trained language models and total parameters.
Compared to previous discriminative systems, we report gains of +$2.2$ CoNLL-F1 points over LingMess, the best encoder-only model. 
Interestingly, we even outperform CorefQA, which uses additional Question Answering training data.

Concerning Sequence-to-Sequence approaches, we report extensive improvements over systems with a similar amount of parameters compared to our large models (500M): we obtain +3.4 points compared to ASP (770M), and the gap is even wider when taking into consideration Link-Append (3B) and seq2seq (770M), with +6.4 and +5.6, respectively.
Most importantly, \sys models surpass the performance of all Sequence-to-Sequence transformers even when they have several billions of parameters.
Among our proposed methods, \sysmes shows the best performance, setting a new state of the art with a score of 83.6 CoNLL-F1 points on the OntoNotes benchmark.
More detailed results, including a table with MUC, B$^3$, and CEAF$\phi_4$ scores and a qualitative error analysis, can be found in Appendix \ref{app:additional-ontonotes-results}. 

\subsection{PreCo and LitBank}\label{sec:results-PreCo-LitBank}
We further validate the robustness of the \sys framework by training and evaluating systems on the PreCo and LitBank datasets.
As reported in Table \ref{table:PreCo-LitBank-training}, our models show superior performance when dealing with long documents in a data-scarce setting such as the one LitBank poses.
On this dataset, \sysincremental achieves a new state-of-the-art score of 78.3, and gains +1.0 CoNLL-F1 points compared with seq2seq.
On PreCo, \sysincremental outperforms longdoc, but seq2seq still shows slightly better performance.
This is mainly due to the high presence of singletons in PreCo (52$\%$ of all the clusters).
Our systems, using a mention extraction technique that favors precision rather than recall, are penalized compared to high recall systems such as seq2seq.\footnote{Precision and Recall scores are reported in Appendix \ref{app:additional-ontonotes-results}.}
Among our systems, \sysincrementalns, leveraging its hybrid architecture, performs better on both PreCo and LitBank. 

\begin{table}[t]
\centering
\small
\resizebox{\columnwidth}{!}{
\begin{tabular}{lcc}
\hline
\multicolumn{1}{c}{Model} & \multicolumn{1}{c}{PreCo}  & \multicolumn{1}{c}{LitBank}\\ 
 \hline
longdoc \cite{toshniwal-etal-2021-generalization} & 87.8 & 77.2 \\
seq2seq \cite{zhang2023seq2seq}& \textbf{88.5} & 77.3 \\ 
\sysstarttoend & 87.2 & 77.6 \\ 
\sysincremental &  88.0 & \textbf{78.3} \\ 
\sysmes &  87.4 & 78.0 \\ 
\hline
\end{tabular}
}
\caption{Results of the compared systems on the PreCo and LitBank test-sets in terms of CoNLL-F1 score.}
\label{table:PreCo-LitBank-training}
\end{table}




\subsection{Out-of-Domain Evaluation}\label{sec:results-ood-evaluation}
In Table \ref{table:ood}, we report the performance of \sys systems along with LingMess, the best encoder-only model, when dealing with out-of-domain texts, that is, when they are trained on OntoNotes and tested on other datasets.
First of all, we report considerable improvements on the GAP test set, obtaining a +1.2 F1 score compared to the previous state of the art. 
We also test models on WikiCoref, PreCo$_{\text{ns}}$ and LitBank$_{\text{ns}}$ (Section \ref{sec:setup-datasets}).
However, since the span annotation guidelines of these corpora differ from the ones used in OntoNotes, in Table \ref{table:ood} we also report the performance using gold mentions, i.e., skipping the mention extraction step (gold column).\footnote{We do not include autoregressive models because none of the original articles report scores on out-of-domain datasets. We also could not test those models because they do not provide the code to perform mention clustering alone, and performing it with such approaches is not as straightforward as in encoder-only models.}
On the WikiCoref benchmark, we achieve a new state-of-the-art score of 67.2 CoNLL-F1, with an improvement of +4.2 points over the previous best score obtained by LingMess. 
On the same dataset, when using pre-identified mentions the gap increases to +5.8 CoNLL-F1 points (76.6 vs 82.4).
In the same setting, our models obtain up to +7.3 and +10.1 CoNLL-F1 points on Preco$_{\text{ns}}$ and LitBank$_{\text{ns}}$, respectively,  compared to LingMess. 
These results suggest that the \sys training strategy makes this model more suitable when dealing with pre-identified mentions and out-of-domain texts.
This further increases the potential benefits that \sys systems can bring to many downstream applications that exploit coreference as an intermediate layer, such as Entity Linking \cite{rosales2020fine} and Relation Extraction \cite{
xiong2023dialogre, zeng2023document}, where the mentions are already identified.
Among our models, on LitBank$_{\text{ns}}$ and WikiCoref, \sysincremental outperforms \sysmes and \sysstarttoendns, confirming the superior capabilities of the incremental formulation in the long-document setting.
Finally, we highlight that the performance gap between using gold mentions and performing full Coreference Resolution is wider when tested on out-of-domain datasets (on average +17$\%$) compared to testing it directly on OntoNotes (83.6 vs 93.6, +10$\%$).\footnote{An evaluation of the proposed \sys models in terms of mention extraction and mention clustering using gold mentions scores can be found in Appendix \ref{app:additional-ontonotes-results}.}
This result, obtained on three different out-of-domain datasets, suggests that the difference in annotation guidelines considerably contribute to lower the OOD performances (-7$\%$).

\begin{table}[t]
\centering
\resizebox{\columnwidth}{!}{
\begin{tabular}{l|c|cc|cc|cc}
\hline
\multicolumn{1}{c}{Model} & \multicolumn{1}{c}
{GAP}  & \multicolumn{2}{c}
{WikiCoref} & \multicolumn{2}{c}{PreCo$_{\text{ns}}$}  & \multicolumn{2}{c}{LitBank$_{\text{ns}}$} \\
\hline
\multicolumn{1}{c}{} & &  sys. & gold & sys. & gold & sys. & gold  \\

 \hline
LingMess &89.6 &  63.0 & 76.6 & 65.1 & 80.6  &  64.4 & 73.9 \\
\sysstarttoend & 91.1 & \textbf{67.2} & 81.5 &   \textbf{67.2} & \textbf{87.9}  & 64.8 & 83.1\\ 
\sysincremental & \textbf{91.2} & 66.8  & \textbf{82.4}  & 66.1 & 86.5  &\textbf{65.4} & \textbf{84.0}\\
\sysmes & 91.1 & 66.8 & 82.1  & 66.1 & 86.9  & 65.1 & 82.8  \\ 
\hline
\end{tabular}
}
\caption{Comparison between LingMess and \sys systems on GAP, WikiCoref, PreCo$_{\text{ns}}$ LitBank$_{\text{ns}}$. We report scores using systems prediction (sys.) or passing gold mentions (gold).}
\label{table:ood}
\end{table}

\subsection{Speed and Memory Usage}\label{sec:speed-memory}
In Table \ref{table:ontonotes-parames-performance}, we include details regarding the training time and the hardware used by each comparison system, along with the measurement of the inference time and peak memory usage on OntoNotes the validation set.
Compared to Coarse-to-Fine models, which require 32GB of VRAM, we can train \sys systems under 18GB.
At inference time both \sysmes and \sysstarttoendns, exploiting DeBERTa$_{\text{large}}$, achieve competitive speed and memory consumption compared to wl-coref and s2e-coref.
Furthermore, when adopting DeBERTa$_{base}$, \sysmes proves to be the most efficient approach\footnote{In terms of inference peak memory usage and speed.} among those directly trained on OntoNotes, while, at the same time, attaining performances that are equal to the previous best encoder-only system, LingMess. 
The only system that shows better inference speed is f-coref, but at the cost of lower performance (-3.0).

Compared to the previous Sequence-to-Sequence state-of-the-art approach, Link-Append, we train our models with 175x less memory requirements. 
Comparing inference time is more complicated, since we could not run models on our memory-constrained budget. 
For this reason, we report the inference times from the original articles, and hence times achieved with their high-resource settings.
Interestingly, we report as much as 170x faster inference compared to seq2seq, which exploits parallel inference on multiple GPUs, and 85x faster when compared to the more efficient ASP.
Among \sys models, \sysincremental is notably slower both in inference and training time, as it incrementally builds clusters using multiple steps.
\begin{table}[t!]
\tiny
\centering
\resizebox{\columnwidth}{!}{
\begin{tabular}{l|l|c}
\hline
\multicolumn{1}{c}{Model} & \multicolumn{1}{c}{LM} & \multicolumn{1}{c}{Score}\\ 
 \hline
\rowcolor{gray!20}
\multicolumn{3}{c}{\textbf{\sysstarttoend}} \\ 
\sysstarttoend & DeBERTa$_{\text{base}}$ & 81.0 \\
 s2e-coref$_{\text{t}}$ & DeBERTa$_{\text{base}}$ & 78.3 \\
 \sysstarttoend & LongFormer$_{\text{large}}$ & 80.6\\
 s2e-coref & LongFormer$_{\text{large}}$ & 80.3 \\
\hline
\rowcolor{gray!20}
\multicolumn{3}{c}{\textbf{\sysmes}} \\ 
 \sysmes  & DeBERTa$_{\text{base}}$ & 81.4\\
 LingMess$_{\text{t}}$  & DeBERTa$_{\text{base}}$ & 78.6 \\
 \sysmes & LongFormer$_{\text{large}}$ & 81.0\\
 LingMess & LongFormer$_{\text{large}}$ & 81.4\\
 \hline
\rowcolor{gray!20}
\multicolumn{3}{c}{\textbf{\sysincremental}} \\ 
 \sysincremental & DeBERTa$_{\text{large}}$ & 83.5\\
 \sysns$_{\text{prev-incr}}$ & DeBERTa$_{\text{large}}$ & 79.6\\
 \hline
\end{tabular}
}
\caption{Comparison between \sys models and previous techniques. LingMess$_{\text{t}}$ and s2e-coref$_{\text{t}}$ are trained using their official scripts. 
We use DeBERTa$_{\text{base}}$ because the DeBERTa$_{\text{large}}$ could not fit our hardware when training comparison systems.}
\label{table:comparison}
\end{table}

\subsection{\sys Ablation}\label{sec:results-comparison-discriminative-models}
In Table \ref{table:comparison}, we compare \sysstarttoend and \sysmes models with s2e-coref and LingMess, respectively, using different pre-trained encoders.
Interestingly, when using DeBERTa, \sys systems not only achieve better speed and memory efficiency, but also obtain higher performance compared to the previous systems. 
When using the LongFormer, instead, their scores are in the same ballpark, showing empirically that the \sys training procedure better exploits the capabilities of DeBERTa.
To test the benefits of our novel incremental formulation, \sysincrementalns, we also implement a \sys model with the previously adopted incremental method used in longdoc and ICoref (Section \ref{sec:related-works-discriminative}), which we call \sysns$_{\text{prev-incr}}$. 
Compared to the previous formulation we report an increase in score of +3.9 CoNLL-F1 points.
The improvement demonstrates that exploiting a transformer architecture to attend to all the previously clustered mentions is beneficial, and enables the future usage of hybrid architectures when needed.

As a further analysis of whether the efficiency improvements of our systems stem from using DeBERTa or are attributable to the \syspipens, we compared the speed and memory occupation of a \sys system using as underlying encoder either DeBERTa$_{\text{large}}$ or LongFormer$_{\text{large}}$.
Our experiments show that using DeBERTa leads to an increase of +77$\%$ of memory space and +23$\%$ of time to complete an epoch when training on OntoNotes. An equivalent measurement, attributable to the quadratic memory attention mechanism of DeBERTa, was observed for the inference time and memory occupation on the OntoNotes test set. These results highlight the efficiency contribution of the \syspipens, which is agnostic to the document encoder and can be applied to future coreference systems to ensure higher efficiency.

\section{Conclusion}\label{sec:conclusions}
In this work, we challenged the recent trends of adopting large autoregressive generative models to solve the Coreference Resolution task. 
To do so, we proposed \sysns, a new framework that enables fast and memory-efficient Coreference Resolution while obtaining state-of-the-art results. 
This demonstrates that the large computational overhead required by Sequence-to-Sequence approaches is unnecessary. 
Indeed, in our experiments \sys systems demonstrated that they can outperform large generative models and improve the speed and memory usage of previous best-performing encoder-only approaches.
Furthermore, we introduced \sysincrementalns, a robust multi-step incremental technique that obtains higher performance in the out-of-domain and long-document setting.
By releasing our systems, we make state-of-the-art models usable by a larger portion of users in different scenarios and potentially improve downstream applications.

\section{Limitations}
Our experiments were limited by our resource setting i.e., a single RTX 4090. For this reason, we could not run \sys using larger encoders, and could not properly test Sequence-to-Sequence models as we did with encoder-only models. Nevertheless, we believe this limitation is a common scenario in many real-world applications that would benefit substantially from our system.
We also did not test our formulation on multiple languages, but note that both the methodology behind \sys and our novel incremental formulation are language agnostic, and thus could be applied to any language. 

\section*{Acknowledgements}
\begin{center}
\noindent
    \begin{minipage}{0.1\linewidth}
        \begin{center}
            \includegraphics[scale=0.05]{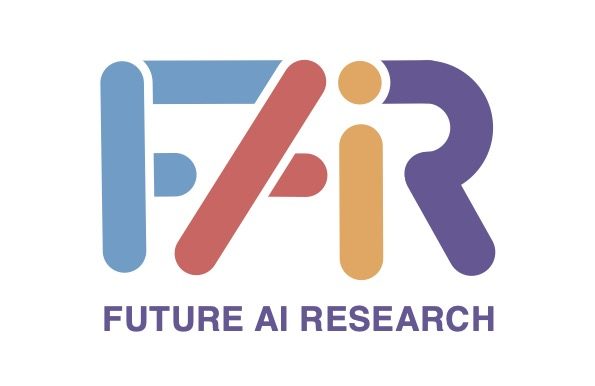}
        \end{center}
    \end{minipage}
    \hspace{0.01\linewidth}
    \begin{minipage}{0.70\linewidth}
         We gratefully acknowledge the support of the PNRR MUR project PE0000013-FAIR.
    \end{minipage}
    \hspace{0.01\linewidth}
    \begin{minipage}{0.1\linewidth}
        \begin{center}
            \includegraphics[scale=0.08]{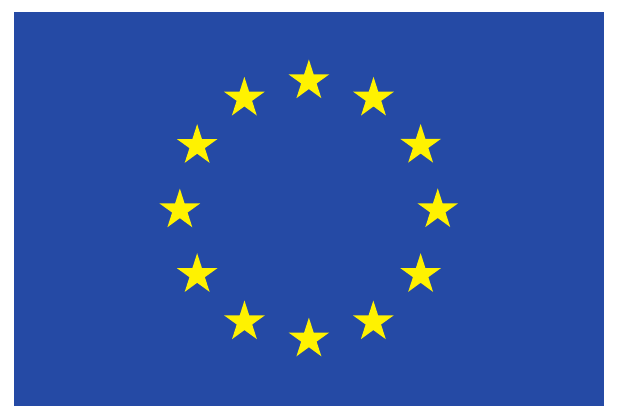}
        \end{center}
    \end{minipage}\\
\end{center}
\vspace{0.2cm}
\noindent Roberto Navigli also gratefully acknowledges the support of the CREATIVE project (CRoss-modal understanding and gEnerATIon of Visual and tExtual content), which is funded by the MUR Progetti di Rilevante Interesse Nazionale programme (PRIN 2020).
This work has been carried out while Giuliano Martinelli was enrolled in the Italian National Doctorate on Artificial Intelligence run by Sapienza University of Rome.

\bibliography{anthology,custom}

\appendix

\section{Multi-Expert Scorers}
\label{app:lingmess}
In \sysmesns, the final coreference score between two spans is calculated using 6 linguistically motivated multi-expert scorers.
This approach was introduced by \citet{otmazgin-etal-2023-lingmess}, which demonstrated that linguistic knowledge and symbolic
computation can still be used to improve results on the OntoNotes benchmark.
In \sysmes we adopt this approach on top of the \syspipens.
We use the same set of categories, namely:
\begin{enumerate}
    \item \textsc{Pron-Pron-C}. Compatible pronouns based on their attributes, such as gender or number (e.g.  \textit{(I, I), (I, my) (she, her)}).  
    \item \textsc{Pron-Pron-NC}, Incompatible pronouns (e.g. \textit{(I, he), (She, my), (his, her)}).
    \item \textsc{Ent-Pron}. Pronoun
    and non-pronoun (e.g. \textit{(George, he), (CNN, it), (Tom Cruise, his)}).
    \item \textsc{Match}. Non-pronoun spans with the same content words (e.g. \textit{Italy, Italy}).
    \item \textsc{Contains}. One contains the other (e.g. \textit{(Barack Obama, Obama)}).
    \item \textsc{Other}. The Other pairs.
\end{enumerate}
To detect pronouns we use string match with a full list of English pronouns. 

To perform mention clustering, we dedicate a mention-pair scorer for each of those categories.
Specifically, for the mention $m_i=(x_s, x_e)$ and its antecedent $m_j=(x_{s'},x_{e'})$, with their start and end token hidden states, we first detect their category $k_g$ using pattern matching on their spans of texts. 
Then we compute their start and end representations, using the specific fully-connected layers for the category $k_g$:
 \begin{equation*}F_s^{k_g}(x) = W_{k_{g,s}}'(GeLU(W_{k_{g,s}}x))
\end{equation*}
 \begin{equation*}F_e^{k_g}(x) = W_{k_{g,e}}'(GeLU(W_{k_{g,e}}x)) 
\end{equation*}
The probability $p_c^{k_g}$ of $m_i$ and $m_j$is then calculated as:
\begin{align*}
p_c^{k_g}(m_i, m_j) = \sigma (F_s^{k_g}(x_s) \cdot  W_{ss} \cdot  F_s^{k_g}(x_{s'}) + \hspace{10mm}\\
F_e^{k_g}(x_e) \cdot W_{ee} \cdot F_e^{k_g}(x_{e'}) +  \hspace{10mm}\\
F_s^{k_g}(x_s) \cdot W_{se} \cdot F_e^{k_g}(x_{e'}) + \hspace{10mm} \\ 
F_e^{k_g}(x_e) \cdot W_{es} \cdot F_s^{k_g}(x_{s'}))\hspace{12mm}
\end{align*}
With $W_{ss}, W_{ee}, W_{se}, W_{es}$ being four learnable matrices and $W_{k_{g,e}}', W_{k_{g,s}}',W_{k_{g,e}}, W_{k_{g,s}}'$ the learnable parameters of the two fully-connected layers.
In this way, each mention-pair scorer learns to model the probability for its specific linguistic categories.

\section{Training details}\label{app:training}
\subsection{Datasets}

We report technical details of the adopted datasets.
\begin{itemize}
    \item \textbf{OntoNotes} contains several items of metadata information for each document, such as genre, speakers, and constituent graphs.
    Following previous works, we incorporate the speaker’s name into the text whenever there is a change in speakers for datasets that include this metadata.
    \item \textbf{LitBank} contains 100 literary documents and is available in 10 different cross-validation folds. Our train, dev, and test splits refer to the first cross-validation fold, LB$_0$. We report comparison systems results on the same splits. Since training DeBERTa$_{\text{large}}$ is particularly computationally expensive, as introduced in Section \ref{sec:setup-sys}, we train on LitBank by splitting in half each LitBank training document.
    \item The authors of \textbf{PreCo} have not released their official test set. 
    To evaluate our models consistently with previous approaches, we use the official 'dev' split as our test set and retain the last 500 training examples for model validation.
\end{itemize}

\begin{table*}[ht!]
\centering
\resizebox{\textwidth}{!}{
\begin{tabular}{l|c|ccc|ccc|ccc|c}
\hline
\multicolumn{1}{c}{Model} & \multicolumn{1}{c}{LM}  & \multicolumn{3}{c}{MUC} & \multicolumn{3}{c}{B$^3$} & \multicolumn{3}{c}{CEAF$\phi_4$} & \multicolumn{1}{c}{Avg.} \\
 & & P & R & F1 & P & R & F1 & P & R & F1 & F1 \\ 
\rowcolor{gray!20}
\hline
\multicolumn{12}{c}{\textbf{Discriminative}} \\ 
e2e-coref \cite{lee-etal-2017-end} & - & 78.4 & 73.4 & 75.8 & 68.6 & 61.8 & 65.0 & 62.7 & 59.0 & 60.8 & 67.2 \\
c2f-coref \cite{lee-etal-2018-higher} & ELMo & 81.4 & 79.5 & 80.4 & 72.2 & 69.5 & 70.8 & 68.2 & 67.1 & 67.6 & 73.0 \\
c2f-coref \cite{joshi-etal-2019-bert} & BERT$_{\text{large}}$ & 84.7 & 82.4 & 83.5 & 76.5 & 74.0 & 75.3 & 74.1 & 69.8 & 71.9 & 76.9 \\
c2f-coref \cite{joshi-etal-2020-spanbert} & SpanBERT$_{\text{large}}$ & 85.8 & 84.8 & 85.3 & 78.3 & 77.9 & 78.1 & 76.4 & 74.2 & 75.3 & 79.6 \\
ICoref \cite{xia-etal-2020-incremental} & SpanBERT$_{\text{large}}$ & 85.7 & 84.8 & 85.3 & 78.1 & 77.5 & 77.8 & 76.3 & 74.1 & 75.2 & 79.4 \\
CorefQA \cite{wu-etal-2020-corefqa} & SpanBERT$_{\text{large}}$ & 88.6 & 87.4 & 88.0 & 82.4 & 82.0 & 82.2 & 79.9 & 78.3 & 79.1 & 83.1* \\
longdoc \cite{toshniwal-etal-2021-generalization} & LongFormer$_{\text{large}}$ & 85.5 & 85.1 & 85.3 & 78.7 & 77.3 & 78.0 & 74.2 & 76.5 & 75.3 & 79.6 \\
s2e-coref Kirstain et al. (2021)
& LongFormer$_{\text{large}}$ & 86.5 & 85.1 & 85.8 & 80.3 & 77.9 & 79.1 & 76.8 & 75.4 & 76.1 & 80.3 \\
wl-coref \cite{dobrovolskii-2021-word} & RoBERTa$_{\text{large}}$ & 84.9 & 87.9 & 86.3 & 77.4 & 82.6 & 79.9 & 76.1 & 77.1 & 76.6 & 81.0 \\
f-coref \cite{otmazgin-etal-2022-f} & DistilRoberta & 85.0 & 83.9 & 84.4 & 77.6 & 75.5 & 76.6 & 74.7 & 74.3 & 74.5 & 78.5* \\
LingMess \cite{otmazgin-etal-2023-lingmess} & LongFormer$_{\text{large}}$ & 88.1 & 85.1 & 86.6 & 82.7 & 78.3 & 80.5 & 78.5 & 76.0 & 77.3 & 81.4 \\
\rowcolor{gray!20}
\hline
\multicolumn{12}{c}{\textbf{Sequence-to-Sequence}} \\ 
TANL \cite{paolini2021structured} & T5$_{\text{base}}$ & - & - & 81.0 & - & - & 69.0 & - & - & 68.4 & 72.8 \\
ASP \cite{liu-etal-2022-autoregressive} & FLAN-T5$_{\text{XXL}}$ & 86.1 & 88.4 & 87.2 & 80.2 & 83.2 & 81.7 & 78.9 & 78.3 & 78.6 & 82.5 \\
Link-Append \cite{bohnet-etal-2023-coreference}  & mT5$_{\text{XXL}}$ & 87.4 & 88.3 & 87.8 & 81.8 & 83.4 & 82.6 & 79.1 & 79.9 & 79.5 & 83.3 \\ 
seq2seq \cite{zhang2023seq2seq}& T0$_{\text{XXL}}$ &86.1 & 89.2 & 87.6 & 80.6 & 84.3 & 82.4 & 78.9 & 80.1 & 79.5 & 83.2 \\ 
\rowcolor{gray!20}
\hline
\multicolumn{12}{c}{\textbf{Ours (Discriminative)}} \\ 
\sysstarttoend &  DeBERTa$_{\text{large}}$ & 87.1 & 88.6 & 87.9 & 81.7 & 83.8 & 82.7 & 80.8 & 78.7 & 79.7 & 83.4\\ 
\sysincremental &  DeBERTa$_{\text{large}}$ & 87.6 & 88.1 & 87.9 & 82.7 & 82.6 & 82.7 & 80.3 & 79.3 & 79.8 & 83.5\\ 
\sysmes &  DeBERTa$_{\text{large}}$ & 87.5 & 88.5 & 88.0 & 82.2 & 83.5 & 82.8 & 80.4 & 79.3 & 79.9 & \textbf{83.6}\\ 
\hline
\end{tabular}
}
\caption{Results on the OntoNotes test set. The average CoNLL-F1 score of MUC, B$^3$, and CEAF$\phi_4$ is the main evaluation criterion. $*$ marks models using additional/different training data.}
\label{table:ontonotes}
\end{table*}

\subsection{Setup}
All our experiments are developed using the Pytorch-Lightning framework.\footnote{\scriptsize\url{https://lightning.ai}}
For each \sys model, we load the pre-trained weights for the \textit{base}\footnote{\scriptsize \url{https://huggingface.co/microsoft/deberta-v3-base}} and \textit{large}\footnote{\scriptsize \url{https://huggingface.co/microsoft/deberta-v3-large}} version of DeBERTa$-$v3 from the Huggingface Transformers library \cite{wolf-etal-2020-transformers}.
We accumulate gradients every 4 steps and use a gradient clipping value of 1.0. 
We adopt a linear learning rate scheduler a warm-up of 10\% of the total steps check validation scores every 50\% of the total number of steps per epoch. 
We select our model upon validation of Avg. CoNLL-f1 score and use patience of 20.

\section{Additional Results}
\label{app:additional-ontonotes-results}
In Table \ref{table:ontonotes} we report the performance of models according to the standard Coreference Resolution metrics: MUC \cite{vilain-etal-1995-model}, B$^3$\cite{bagga-baldwin-1998-entity-based}, CEAF$\phi_4$ \cite{luo-2005-coreference} and AVG CoNLL-F1.
Scores for \sys models are computed using the official CoNLL coreference scorer.\footnote{\scriptsize\url{https://conll.github.io/reference-coreference-scorers}}

\subsection{Error Analysis}
To better understand the quality of \sys predictions, we conduct an error analysis on our best system trained on OntoNotes, \sysmesns. 
In Table \ref{table:gap}, we report the score of performing only mention extraction (F1) or mention clustering with gold mention (CoNLL-F1) with our systems.
Our results highlight that our models have strong capabilities of clustering pre-identified mentions, but limited performance in the identification of correct spans.
We investigated this phenomenon by conducting a qualitative evaluation of the outputs of our best system, \sysmesns, and found out that OntoNotes contains several annotation errors.
We report examples of errors in Table \ref{table:error-analysis-qualitative}. The main inconsistency we found in the gold test set is that many documents have incomplete annotations, which directly correlates with the mention extraction error.

\begin{table}[t!]
\centering
\resizebox{\columnwidth}{!}{
\begin{tabular}{lcc}
\hline
System & Ment. Clustering & Ment. Extraction\\
\hline
\sysstarttoend & 89.4 & 93.5\\
\sysincremental & 89.2 & 94.2\\
\sysmes & 89.6 & 93.7\\
\hline
\end{tabular}
}
\caption{Mention extraction (F1) and mention clustering (CoNLL-F1) scores on the OntoNotes validation set.}
\label{table:gap}
\end{table}

\begin{table*}[!t]
\centering
\resizebox{\linewidth}{!}{%
\begin{tabular}{l|l}
\hline
\textbf{Type} & \textbf{Text}\\
\hline
Ex. 1 & 
\\
\hline
Gold & Nine people were injured in Gaza when gunmen \abo \textbf{opened}\abc \bbo \textbf{fire}\bbc on an Israeli bus. \\
& The passengers were off - duty Israeli security workers. \\
& Witnesses say \bbo \textbf{the shots}\bbc came from \cbo \textbf{the Palestinian international airport}\cbc. \\
& Israeli Prime Minister Ehud Barak \dbo \textbf{closed}\dbc down \cbo \textbf{the two - year - old airport}\cbc in response to \abo \textbf{the incident}\abc. \\
& \ebo \textbf{Palestinians}\ebc criticized \dbo \textbf{the move}\dbc. \\
& \ebo \textbf{hey}\ebc regard \cbo \textbf{the airport}\cbc as a symbol of emerging statehood.\\

\hline
Output 
& \abo \textbf{Nine people}\abc were injured in Gaza when gunmen opened fire on an Israeli bus. \\ 
& \abo \textbf{The passengers}\abc were off - duty Israeli security workers. \\ 
& Witnesses say the shots came from \bbo \textbf{the Palestinian international airport}\bbc. \\ 
& Israeli Prime Minister Ehud Barak \cbo \textbf{closed}\cbc down \bbo \textbf{the two - year - old airport}\bbc in response to the incident. \\ 
&\dbo \textbf{Palestinians}\dbc criticized \cbo \textbf{the move}\cbc. \\ 
&\dbo \textbf{They}\dbc regard \bbo \textbf{the airport}\bbc as a symbol of emerging statehood.\\
\hline

Ex. 2 & 
\\

\hline
Gold 
& \abo \textbf{Mr. Seelenfreund}\abc is \bbo \textbf{executive vice president and chief financial officer of }\cbo \textbf{McKesson}\cbc\bbc -\\
& and will continue in \bbo \textbf{those roles}\bbc. \\
& \dbo \textbf{PCS}\dbc also named Rex R. Malson, 57, executive vice president at McKesson,- \\
& as a director, filling the seat vacated by Mr. Field. \\
& Messrs. Malson and Seelenfreund are directors of \cbo \textbf{McKesson, which has an 86$\%$ stake in }\dbo \textbf{PCS}\dbc\cbc.\\

\hline
Output 
& \abo \textbf{Mr. Seelenfreund}\abc is \bbo \textbf{executive vice president and chief financial officer of }\cbo \textbf{McKesson}\cbc\bbc \\
& and will continue in \bbo those roles\bbc. \\
& \dbo \textbf{PCS}\dbc  also named \ebo \textbf{Rex R. Malson, 57, executive vice president at } \cbo \textbf{McKesson}\cbc \textbf{,}\ebc - \\
& as a director, filling the seat vacated by Mr. Field. \\
& Messrs. \ebo \textbf{Malson}\ebc and \abo \textbf{Seelenfreund}\abc are directors of \cbo\textbf{McKesson, which has an 86 $\%$ stake in }\dbo \textbf{PCS}\dbc\cbc.\\

\hline
Ex. 3 & 
\\

\hline
Gold 
& The Second U.S. Circuit Court of Appeals opinion in the Arcadian Phosphate case  - \\
& did not repudiate the position \abo \textbf{Pennzoil Co.}\abc took in \abo \textbf{its}\abc  dispute with \bbo \textbf{Texaco}\bbc, -\\
& contrary to your Sept. 8 article `` Court Backs \bbo \textbf{Texaco}\bbc 's View in \abo \textbf{Pennzoil}\abc  Case -- Too Late. '' \\
& The fundamental rule of contract law applied to \cbo \textbf{both cases}\cbc was that courts will not enforce  - \\
& \dbo \textbf{agreements to } \dbo \textbf{which}\dbc \textbf{the parties did not intend to be bound}\dbc.\\ 
& In the Pennzoil / Texaco litigation, \ebo \textbf{the courts}\ebc found \abo \textbf{Pennzoil}\abc  and Getty Oil intended to be bound; \\
& in Arcadian Phosphates \ebo \textbf{they}\ebc found there was no intention to be bound. \\

\hline
Output 
& The Second U.S. Circuit Court of Appeals opinion in \abo \textbf{the Arcadian Phosphate case}\abc\\
& - did not repudiate the position \bbo \textbf{Pennzoil Co.}\bbc took in 
\cbo\cbo\bbo \textbf{its}\bbc \textbf{dispute with }\dbo \textbf{Texaco}\dbc\cbc\textbf{, -} \\ 
& \textbf{contrary to your Sept. 8 article `` Court Backs }\dbo \textbf{Texaco 's}\dbc \textbf{View in }\cbo \bbo \textbf{Pennzoil}\bbc \textbf{Case}\cbc\cbc -- Too Late . '' \\
& \ebo\ebo \textbf{The fundamental rule of contract law}\ebc  \textbf{applied to both cases}\ebc was that courts will not enforce - \\
& agreements to which the parties did not intend to be bound. \\
& In \cbo \textbf{the} \bbo \textbf{Pennzoil}\bbc \textbf{ /} \dbo \textbf{Texaco}\dbc \textbf{litigation}\cbc, \fbo \textbf{the courts}\fbc  found \bbo \textbf{Pennzoil}\bbc and Getty Oil intended to be bound; \\
& in \abo \textbf{Arcadian Phosphates}\abc \fbo \textbf{they}\fbc found there was no intention to be bound. \\

\hline
Ex. 4 & 
\\
\hline
Gold 
&... \abo \textbf{Harry}\abc has avoided all that by living in a Long Island suburb with \abo \textbf{his}\abc wife, \\
& who 's so addicted to soap operas and mystery novels \\
& she barely seems to notice when \abo \textbf{her husband}\abo disappears for drug - seeking forays into Manhattan.\\

\hline
Output 
&... \abo \textbf{Harry}\abc has avoided all that by living in a Long Island suburb with \bbo \abo \textbf{his}\abc \textbf{wife, }\\
& \textbf{who 's so addicted to soap operas and mystery novels}\\
& \bbo \textbf{she}\bbc \textbf{barely seems to notice when }\abo \bbo \textbf{her}\bbc \textbf{husband}\abc \textbf{disappears for drug - seeking forays into Manhattan}\bbc. \\

\hline

\end{tabular}
}
\caption{OntoNotes test set annotation errors examples.}
\label{table:error-analysis-qualitative}
\end{table*}

\end{document}